\documentclass[letterpaper, 10 pt, conference, twoside]{IEEEtran}  %
\IEEEoverridecommandlockouts                              %

\usepackage{mathtools}
\usepackage{graphicx,subcaption} %
\graphicspath{{img/}}
\usepackage{pgf}
\usepackage{calc}
\newcommand\inputpgf[2]{{
\let\pgfimageWithoutPath\pgfimage
\renewcommand{\pgfimage}[2][]{\pgfimageWithoutPath[##1]{#1/##2}}
\input{#1/#2}
}}

\usepackage{todonotes}
\usepackage{soul}

\usepackage{etoolbox}

\usepackage{amsmath}

\usepackage{amssymb}
\usepackage{bm}
\usepackage[export]{adjustbox}
\usepackage{tabularx}
\newcolumntype{Y}{>{\centering\arraybackslash}X} %

\usepackage{tikz}
\usetikzlibrary{bayesnet} %

\newlength{\state} %
\usepackage{url}
\usepackage{array}
\usepackage{textcomp}
\usepackage{xspace}
\usepackage{booktabs}
\usepackage{tablefootnote}

\usepackage{cite}

\usepackage{color}
\newcommand{\blue}[1]{{\color{blue}#1}}

\usepackage{hyperref}
\hypersetup{
    colorlinks,
    linkcolor={red!50!black},
    citecolor={blue!50!black},
    urlcolor={blue!80!black}
}

\usepackage[capitalize]{cleveref}
\renewcommand{\Cref}[1]{\cref{#1}} %

\newcommand{\linkToPdf}[1]{\href{#1}{\blue{(pdf)}}}
\newcommand{\linkToPpt}[1]{\href{#1}{\blue{(ppt)}}}
\newcommand{\linkToCode}[1]{\href{#1}{\blue{(code)}}}
\newcommand{\linkToWeb}[1]{\href{#1}{\blue{(web)}}}
\newcommand{\linkToVideo}[1]{\href{#1}{\blue{(video)}}}
\newcommand{\award}[1]{\xspace} %

\newcommand{\mysubsection}[1]{{\bf#1.}}

\title{Incremental Visual-Inertial 3D Mesh \\ Generation with Structural Regularities}

\author{Antoni Rosinol$^{1}$, Torsten Sattler$^{2}$, Marc Pollefeys$^{3}$, Luca Carlone$^{1}$
\thanks{$^{1}$A.\,Rosinol and L.\,Carlone are with the Laboratory for
Information \& Decision Systems (LIDS), Massachusetts Institute of Technology, Cambridge, MA, USA,
{\sf \{arosinol,lcarlone\}@mit.edu}}
\thanks{$^{2}$T.\,Sattler is with the Department of Electrical Engineering, Chalmers University of
Technology, Sweden. This work was done while Torsten was at ETH Z\"urich, {\sf torsat@chalmers.se}}
\thanks{$^{3}$M.\,Pollefeys is with the Department of Computer Science, ETH Z\"urich, and with
Microsoft, Switzerland, {\sf marc.pollefeys@inf.ethz.ch}}
\thanks{This work was partially funded by ARL DCIST CRA W911NF-17-2-0181, Lincoln Laboratory, and the Zeno Karl Schindler foundation.}}%

\newcommand{\hide}[1]{}

\newcommand{\bdmath}{\begin{dmath}}%
\newcommand{\edmath}{\end{dmath}}
\newcommand{\beq}{\begin{equation}}
\newcommand{\eeq}{\end{equation}}
\newcommand{\bdm}{\begin{displaymath}}
\newcommand{\edm}{\end{displaymath}}
\newcommand{\bea}{\begin{eqnarray}}
\newcommand{\eea}{\end{eqnarray}}
\newcommand{\beal}{\beq \begin{array}{ll}}
\newcommand{\eeal}{\end{array} \eeq}
\newcommand{\beas}{\begin{eqnarray*}}
\newcommand{\eeas}{\end{eqnarray*}}
\newcommand{\ba}{\begin{array}} %
\newcommand{\ea}{\end{array}}

\newcommand{\bit}{\begin{itemize}}
\newcommand{\eit}{\end{itemize}}
\newcommand{\ben}{\begin{enumerate}}
\newcommand{\een}{\end{enumerate}}
\newcommand{\bbmat}{\begin{bmatrix}}
\newcommand{\ebmat}{\end{bmatrix}}
\newcommand{\bpmat}{\begin{pmatrix}}
\newcommand{\epmat}{\end{pmatrix}}

\newcommand{\SO}{\mathrm{SO}}

\newcommand{\Sphere}{\mathrm{S}}

\newcommand{\Real}{\mathbb{R}}

\newcommand{\SOthree}{\ensuremath{\SO(3)}\xspace}

\newcommand{\Stwo}{\ensuremath{\Sphere^2}\xspace}

\newcommand{\Rthree}{\ensuremath{\mathbb{R}^3}\xspace}

\newcommand{\calK}{{\cal K}}
\newcommand{\calL}{{\cal L}}

\newcommand{\calR}{{\cal R}}
\newcommand{\calS}{{\cal S}}

\newcommand{\calX}{{\cal X}}

\newcommand{\R}{\mathbf{R}}

\newcommand{\residual}{\mathbf{r}}
\newcommand{\transpose}{\mathsf{T}} %

\newcommand{\tran}{\mathbf{p}}

\newcommand{\vel}{\mathbf{v}}

\newcommand{\bias}{\mathbf{b}}

\newcommand{\LandmarkSet}{\ensuremath{\calL}\xspace}

\newcommand{\Plane}{\ensuremath{\boldsymbol{\pi}}\xspace}
\newcommand{\PlaneSet}{\ensuremath{\Pi}\xspace}

\newcommand{\Normal}{\ensuremath{\boldsymbol{n}}\xspace}

\newcommand{\normsq}[2]{\left\|#1\right\|^2_{#2}}

\newcommand{\meascam}{\mathcal{C}}
\newcommand{\measimu}{\mathcal{I}}

\newcommand{\landmark}{\bm{\rho}}

\newcommand{\subimu}{{ij}}
\newcommand{\indmeas}{(i,j)}

\newcommand{\TimeWindow}{\ensuremath{\Delta_t}\xspace}

\newcommand{\Euroc}{EuRoC\xspace}

\begin{document}

\maketitle

\begin{tikzpicture}[overlay, remember picture]
\path (current page.north east) ++(-6.3,-0.2) node[below left] {
Accepted for publication at ICRA 2019, please cite as follows:
};
\end{tikzpicture}
\begin{tikzpicture}[overlay, remember picture]
\path (current page.north east) ++(-7.2,-0.6) node[below left] {
A. Rosinol, T. Sattler, M. Pollefeys, L. Carlone
};
\end{tikzpicture}
\begin{tikzpicture}[overlay, remember picture]
\path (current page.north east) ++(-4.6,-1) node[below left] {
``Incremental Visual-Inertial 3D Mesh  Generation with Structural Regularities'',
};
\end{tikzpicture}
\begin{tikzpicture}[overlay, remember picture]
\path (current page.north east) ++(-7.5,-1.4) node[below left] {
 IEEE Int. Conf. Robot. Autom. (ICRA), 2019.
};
\end{tikzpicture}

\begin{abstract}
 Visual-Inertial Odometry (VIO) algorithms typically rely on a point cloud representation of the scene that does not model the topology of the environment.
 A 3D mesh instead offers a richer, yet lightweight, model.
 Nevertheless, building a 3D mesh out of the sparse and noisy 3D landmarks triangulated by a VIO algorithm often results in a mesh that does not fit the real scene.
  In order to regularize the mesh, previous approaches decouple state estimation from the 3D mesh regularization step, and either limit the 3D mesh to the current frame~\cite{Greene17iccv,Teixeira16iros} or let the mesh grow indefinitely~\cite{Pollefeys2008ijcv,Litvinov2013bmvc}.
 We propose instead to tightly couple mesh regularization and state estimation by detecting and enforcing \emph{structural regularities} in a novel factor-graph formulation.
 We also propose to incrementally build the mesh by restricting its extent to the time-horizon of the VIO optimization; the resulting 3D mesh covers a larger portion of the scene than a per-frame approach while its memory usage and computational complexity remain bounded.
We show that our approach successfully regularizes the mesh, while improving localization accuracy, when structural regularities are present,
and remains operational in scenes without regularities.
\end{abstract}


\section*{Supplementary Material}
\href{https://www.mit.edu/~arosinol/research/struct3dmesh.html}{https://www.mit.edu/\texttildelow arosinol/research/struct3dmesh.html}

\section{Introduction}
\label{sec:introduction}

\begin{figure}[t]
  \centering
  \includegraphics[width=\columnwidth]{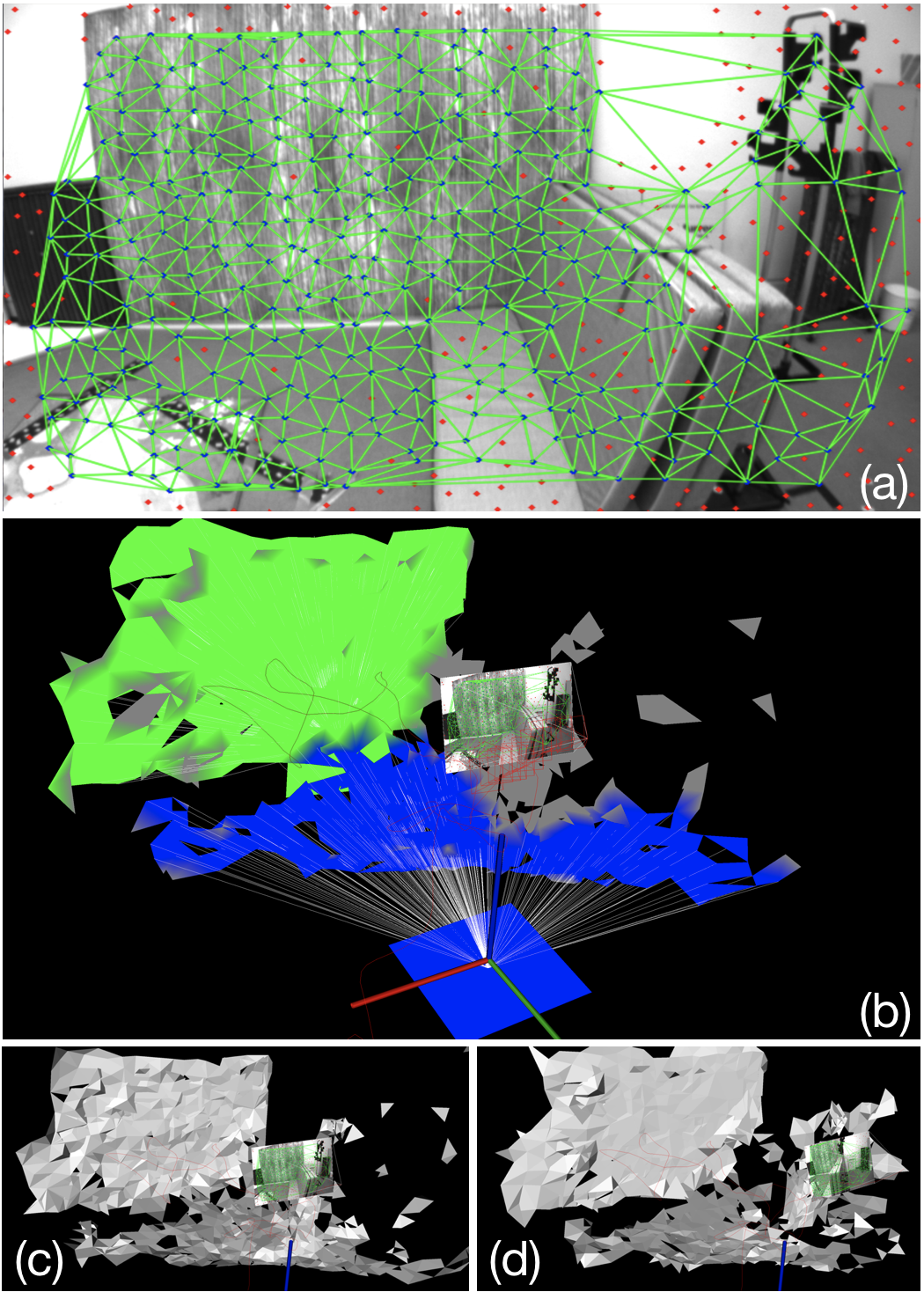}
  \caption{We propose a VIO pipeline that incrementally builds a 3D mesh of the environment starting from a 2D Delaunay triangulation of keypoints (a). We also detect and enforce \emph{structural regularities}, c.f.~(b) planar walls (green) and floor (blue). The bottom row compares the mesh constructed
  (c) without and (d) with structural regularities.\vspace{-1em}}
  \label{fig:intro}
\end{figure}

 Recent advances in VIO are enabling a wide range of applications, ranging from
 virtual and augmented reality to agile drone navigation~\cite{SayreMcCord18icra}.
While VIO methods can deliver accurate state estimates in real-time, they typically provide
a sparse map of the scene.
In particular, feature-based methods~\cite{Mourikis07icra, Leutenegger15ijrr, Forster17troOnmanifold, Mur17} produce a point cloud that is not directly usable for path planning or obstacle avoidance.
In those cases, a denser map is built subsequently, e.g., by using (multi-view) stereo algorithms~\cite{Schoeps2017cvpr, Pillai16icra}.
Alternatively, direct every-pixel methods estimate denser point clouds online~\cite{Newcombe10cvpr, Newcombe11iccv, Izadi11siggraph}.
Nevertheless, these algorithms rely on GPUs which consume relatively high amounts of power, making them impractical for computationally-constrained systems such as micro aerial vehicles or smartphones.
Furthermore, these models typically decouple trajectory estimation and mapping, resulting in a loss of accuracy~\cite{Platinsky17icra}, and produce representations that are expensive to store and manipulate.
  Ideally one would like to use a map representation that (i) is lightweight to compute and store, (ii) describes the topology of the environment, and (iii) couples state estimation and mapping, allowing one to improve the other and vice versa.
  A 3D mesh representation is lightweight, while it provides information about the topology of the scene.
  Moreover, a 3D mesh allows for extracting the structure of the scene, which can potentially be used to improve simultaneously the accuracy of the pose estimates and the mesh itself, thereby coupling state estimation and mapping.

  Recent approaches have tried to avoid the caveats of every-pixel methods by using a 3D mesh over the set of sparse 3D landmarks triangulated by a VIO pipeline.
  Nevertheless, these approaches perform regularization of the mesh as a post-processing step -- decoupling state estimation and mesh generation -- and work on a per-frame basis \cite{Greene17iccv, Teixeira16iros}.
  Our approach instead tightly couples the 3D mesh generation and the state estimation by enforcing structural constraints in a factor-graph formulation, which allows for joint mesh regularization and pose estimation.
  We also maintain the 3D mesh over the receding horizon of the VIO's fixed-lag optimization problem, thereby spanning multiple frames and covering a larger area than the camera's immediate field-of-view.

{\bf Contributions.}
In this paper, we propose to \emph{incrementally build a 3D mesh restricted to the receding horizon of the VIO optimization.}
In this way, we can map larger areas than a per-frame approach, while memory footprint and computational complexity associated to the mesh remain bounded.
We also propose to use the 3D mesh to \emph{detect and enforce structural regularities} in the optimization problem, thereby improving the accuracy of both the state estimation and the mesh at each iteration, while circumventing the need for an extra regularization step for the mesh.
In particular, we extract co-planarity constraints between landmarks (\cref{fig:intro}), and show that we can detect these structural priors in a non-iterative way, contrary to RANSAC-based approaches \cite{li17improved}.
Overall, our approach runs in real-time by using a single \mbox{CPU} core.
Moreover, we do not rely on sensors such as \mbox{LIDAR} or \mbox{RGB-D} cameras, instead we use a (stereo) monochrome camera.

Finally, we provide an extensive experimental evaluation on the \Euroc{} dataset~\cite{Burri15ijrr},
 where we compare the proposed VIO approach against state-of-the-art methods.
Our evaluation shows that (i) the proposed approach produces a lightweight representation of the environment that captures the geometry of the scene,
 (ii)~leveraging structural regularities improves the state and map estimation, surpassing the state-of-the-art when structural regularities are present, while (iii) performing on-par with standard VIO methods in absence of regularities.

\vspace{-0em}
\section{Approach}
\label{sec:mathematical_formulation}
\vspace{-0.1em}

We consider a stereo visual-inertial system and adopt a \emph{keyframe}-based approach~\cite{Leutenegger15ijrr}.
This section describes our VIO front-end and back-end.
Our front-end proceeds by building a 2D Delaunay triangulation over the 2D keypoints at each keyframe.
Then, the VIO back-end estimates the 3D position of each 2D keypoint, which we use to project the 2D triangulation into a 3D mesh.
While we incrementally build the 3D mesh, we restrict the mesh to the time-horizon of the VIO optimization, which we formulate in a fixed-lag smoothing framework~\cite{Qin17arxiv, Carlone17icra-vioAttention}.
The 3D mesh is further used to extract structural regularities in the scene that are then encoded as constraints in the back-end optimization.

\subsection{Front-end}
\label{ssec:frontend}

Our front-end has the same components as a keyframe-based indirect visual-inertial odometry pipeline\cite{Leutenegger15ijrr, Blosch15iros}, but it also incorporates a module to generate a 3D mesh, and a module to detect structural regularities from the 3D mesh.
We refer the reader to \cite[Sec. 4.2.1]{RosinolMT} for details on the standard modules used, and we focus here instead on the 3D mesh generation and regularity detection.

\subsubsection{3D Mesh Generation}
using a sparse point cloud from VIO to create a 3D mesh is difficult because (i) the 3D positions of the landmarks are noisy, and some are outliers; (ii) the density of the point cloud is highly irregular; (iii) the point cloud is constantly morphing: points are being removed (marginalized) and added, while the landmarks' positions are being updated at each optimization step.
Therefore, we avoid performing a 3D tetrahedralisation from the landmarks, which would require expensive algorithms, such as space carving \cite{Pan2009proforma}.
Instead, we perform a 2D Delaunay triangulation only over the tracked keypoints in the latest frame,
 as shown in \cref{fig:intro} (a); and project the 2D triangulation in 3D using the fact that each tracked keypoint has a 3D landmark associated (\cref{fig:intro} (b)).
 For the first frame, no keypoint is yet tracked, hence no 3D mesh is generated.

The Delaunay triangulation maximizes the minimum angle of all the angles of the triangles in the triangulation; thereby avoiding triangles with extremely acute angles.
Since we want to promote triangles that represent planar surfaces, this is a desirable property, as it will promote near isotropic triangles that cover a good extent of a potentially planar surface.
Nevertheless, having an isotropic triangle in 2D does not guarantee that the corresponding triangle in 3D will be isotropic, as one of the vertices could be projected far from the other two.
Furthermore, a triangle in the 2D image will result in a 3D triangle independently of whether it represents an actual surface or not.
We deal with some of these misrepresentative faces of the mesh by using simple geometric filters that we detail in \cite[Sec. 3.2.1]{RosinolMT}.

\subsubsection{3D Mesh Propagation}
While some algorithms update the mesh for a single frame \cite{Greene17iccv,Teixeira16iros}, we attempt to maintain a mesh over the receding horizon of the fixed-lag smoothing optimization problem (\cref{ssec:backend}), which contains multiple frames.
The motivation is three-fold: (i) A mesh spanning multiple frames covers a larger area of the scene, which provides more information than just the immediate field of view of the camera. (ii) We want to capture the structural regularities affecting any landmark in the optimization problem.
(iii) Anchoring the 3D mesh to the time-horizon of the optimization problem also bounds the memory usage, as well as the computational complexity of updating the mesh.
The 3D mesh propagation can be decomposed in two parts.

\textit{a) Temporal propagation} deals with the problem of updating the 3D mesh when new keypoints appear and/or old ones disappear in the new frame.
Unfortunately, most of the keypoints' positions on the 2D image change each time the camera moves.
Hence, we re-compute a 2D Delaunay triangulation from scratch over the keypoints of the current frame.
  We can then project all the 2D triangles to 3D mesh faces, since we are keeping track of the landmark associated to each keypoint.

\textit{b) Spatial propagation} deals with the problem of updating the global 3D mesh when a new local 3D mesh is available, and when old landmarks are marginalized from the optimization's time-horizon.
We solve the first problem by merging the new local 3D mesh to the previous (global) mesh, by ensuring no duplicated 3D faces are present. 
  At the same time, when a landmark is marginalized from the optimization, we remove any face in the 3D mesh that has the landmark as a vertex.
  This operation is not without caveats.
  For example, the removed landmark might be at the center of a wall, thereby leaving a hole when surrounding faces of the mesh are deleted.
  While we did not attempt to solve this issue, the problem usually appears on the portion of the mesh that is not currently visible by the camera.
  Also, we do not explicitly deal with the problem of occlusions.

\subsubsection{Regularity Detection}
\label{sssec:regularity_detection}

By reasoning in terms of the triangular faces of the mesh, we can extract the geometry in the scene in a non-iterative way (unlike RANSAC approaches).
In particular, we are interested in co-planarity regularities between landmarks, for which we need to first find planar surfaces in the scene.
In our approach, we only detect planes that are either vertical (i.e.~walls) or horizontal (i.e.~floor, tables), which are structures commonly found in man-made environments.
\Cref{fig:intro} (b) shows the faces of the mesh associated to a vertical wall in green, while the blue faces correspond to the floor.
To detect horizontal planes, we cluster the faces of the mesh with vertical normals, and then build a 1D histogram of the height of the vertices.
After smoothing the histogram with a Gaussian filter, the resulting local maximums of the histogram correspond to predominant horizontal planes.
Among these planes, we take the candidates with the most inliers (above a minimum threshold of $20$ faces).
To detect vertical planes, we cluster the faces of the mesh which have a horizontal normal.
Then, we build a 2D histogram; where one axis corresponds to the shortest distance from the plane of the 3D face to the world origin\footnote{The world origin corresponds to the first estimated pose of the IMU.},
 and the other axis corresponds to the azimuth of the normal with respect to the vertical direction\footnote{Since gravity is observable via the IMU, we have a good estimate of what the vertical direction is.}.
Candidate selection is done the same way as in the horizontal case.

\subsubsection{Data Association}
\label{sssec:data_association}

With the newly detected planes, we still need to associate which landmarks are on each plane.
For this, we use the set of landmarks of the 3D faces that voted for the given plane in the original histogram.
Once we have a new set of planes detected, we still need to check if these planes are already present in the optimization problem to avoid duplicated plane variables.
For this, we simply compare the normals and distances to the origin of the plane to see if they are close to each other.

\subsection{Back-end}
\label{ssec:backend}

\subsubsection{State Space}
\label{sssec:state_space}

If we denote the set of all keyframes up to time $t$ by $\calK_t$, the state of the system $\mathbf{x}_i$ at keyframe $i\in\calK_t$ is described by the IMU orientation $\R_i \in \SOthree$,
 position $\tran_i \in \Rthree$, velocity $\vel_i \in \Rthree$, and biases $\bias_i = [\bias^g_i \;\; \bias^a_i] \in \Real^6$, where $\bias^g_i, \bias^a_i \in \Real^3$ are respectively the gyroscope and accelerometer biases:
\beq
\mathbf{x}_i \doteq [\R_i,\tran_i,\vel_i,\bias_i].
\eeq

We will only encode in the optimization the 3D positions $\landmark_l$ for a subset $\Lambda_t$ of all landmarks $\LandmarkSet_t$ visible up to time $t$: $\{\landmark_l\}_{l\in\Lambda_t}$, where $\Lambda_t \subseteq \LandmarkSet_t$.
We will avoid encoding the rest of the landmarks $\calS_t = \LandmarkSet_t\setminus\Lambda_t$ by using a structureless approach, as defined in \cite[Sec. VII]{Forster17troOnmanifold}, which circumvents the need to add the landmarks' positions as variables in the optimization.
This allows trading-off accuracy for speed, since the optimization’s complexity increases with the number of variables to be estimated.

The set $\Lambda_t$ corresponds to the landmarks which we consider to satisfy a structural regularity.
In particular, we are interested in co-planarity regularities, which we introduce in \cref{sssec:regularity_constraints}.
Since we need the explicit landmark variables to formulate constraints on them, we avoid using a structureless approach for these landmarks.
Finally, the co-planarity constraints between the landmarks $\Lambda_t$ require the modelling of the planes $\PlaneSet_t$ in the scene.
Therefore, the variables to be estimated comprise the state of the system $\{\mathbf{x}_i\}_{i\in\calK_t}$, the landmarks which we consider to satisfy structural regularities $\{\landmark_l\}_{l\in\Lambda_t}$, and the planes $\{\Plane_\pi\}_{\pi\in\PlaneSet_t}$.
The variables to be estimated at time $t$ are:
\begin{equation}
  \calX_t \doteq \displaystyle\left\{\mathbf{x}_i, \landmark_l, \Plane_\pi\right\}_{i\in\calK_t, l\in\Lambda_t, \pi\in\Pi_t}.
  \label{eq:state_vector}
\end{equation}
Since we are taking a fixed-lag smoothing approach for the optimization, we limit the estimation problem to the sets of variables in a time-horizon of length $\TimeWindow$.
To avoid cluttering the notation, we skip the dependence of the sets $\calK_t$, $\Lambda_t$ and $\Pi_t$ on the parameter $\TimeWindow$.
By reducing the number of variables to a given window of time $\TimeWindow$, we will make the optimization problem more tractable and solvable in real-time.

\subsubsection{Measurements}
\label{sssec:measurements}

The input for our system consists of measurements from the camera and the IMU.
We define the image measurements at keyframe $i$ as $\meascam_i$.
The camera can observe multiple landmarks $l$, hence $\meascam_i$ contains multiple image measurements $\mathbf{z}_{i}^{l}$,
 where we distinguish the landmarks that we will use for further structural regularities $l_c$ (where the index $c$ stands for `constrained' landmark),
  and the landmarks that will remain as structureless $l_s$ (where the index $s$ stands for `structureless').
We represent the set of IMU measurements acquired between two consecutive keyframes $i$ and $j$ as  $\measimu_\subimu$.
Therefore, we define the set of measurements collected up to time $t$ by $\mathcal{Z}_t$:
\begin{equation}
  \mathcal{Z}_t \doteq \{\meascam_i, \measimu_\subimu\}_{\indmeas \in \mathcal{K}_t}.
  \label{eq:measurements}
\end{equation}

\subsubsection{Factor Graph Formulation}
\label{sssec:factor_graph}
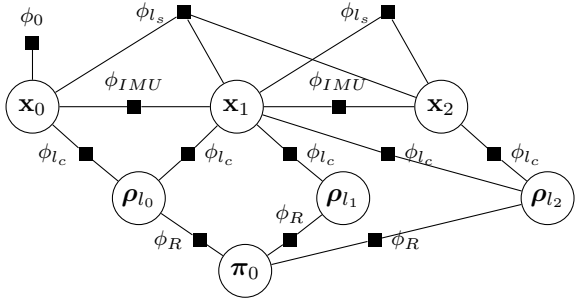
\begin{figure}[h]
  \centering
  \begin{tikzpicture}
    \node[latent] (x0) {$\mathbf{x}_0$};
    \node[latent, right= of x0, xshift=\state] (x1) {$\mathbf{x}_1$};
    \node[latent, right= of x1, xshift=\state] (x2) {$\mathbf{x}_2$};

    \node[latent, below=0.5 of x0, xshift=1.414\state] (l0) {$\bm\rho_{l_0}$};
    \node[latent, below=0.5 of x1, xshift=1.414\state] (l1) {$\bm\rho_{l_1}$};
    \node[latent, below=0.5 of x2, xshift=1.414\state] (l2) {$\bm\rho_{l_2}$};

    \node[latent, below=0.25 of l0, xshift=1.414\state] (pi0) {$\bm\pi_0$};

    \factor[above=of x0, xshift=0\state] {x0prior} {above:$\phi_{0}$} {x0} {}; %

    \factor[right=of x0, xshift=0.5\state] {x0-x1} {above:$\phi_{IMU}$} {x0, x1} {}; %
    \factor[right=of x1, xshift=0.5\state] {x1-x2} {above:$\phi_{IMU}$} {x1, x2} {}; %

    \factor[below=0.18 of x0, xshift=0.707\state] {x0-l0} {left:$\phi_{l_c}$} {x0, l0} {}; %
    \factor[below=0.18 of x1, xshift=-0.650\state] {x1-l0} {right:$\phi_{l_c}$} {x1, l0} {}; %

    \factor[below=0.18 of x1, xshift=0.707\state] {x1-l1} {right:$\phi_{l_c}$} {x1, l1} {}; %

    \factor[below=0.18 of x2, xshift=0.707\state] {x2-l2} {right:$\phi_{l_c}$} {x2, l2} {}; %
    \factor[below=0.18 of x2, xshift=-0.707\state] {x1-l2} {right:$\phi_{l_c}$} {x1, l2} {}; %

    \factor[above=0.8of x1, xshift=-0.707\state] {x0-x1-x2} {left:$\phi_{l_s}$} {x0, x1,x2} {}; %
    \factor[above=0.8of x2, xshift=-0.707\state] {x1-x2} {left:$\phi_{l_s}$} {x1,x2} {}; %

    \factor[below=0.1 of l0, xshift=0.807\state] {pi0-l0-l1} {left:$\phi_{R}$} {l0, pi0} {}; %
    \factor[below=0.1of l1, xshift=-0.707\state] {pi0-l1} {above:$\phi_{R}$} {l1, pi0} {}; %
    \factor[below=0.1of l2, xshift=-2.3\state] {pi0-l2} {right:$\phi_{R}$} {l2, pi0} {}; %
  \end{tikzpicture}
  \caption{VIO factor graph combining Structureless ($\phi_{l_s}$), Projection ($\phi_{l_c}$) and Regularity ($\phi_{R}$) factors (SPR).
The factor $\phi_{R}$ encodes relative constraints between a landmark $l_i$ and a plane $\pi_0$. \vspace{-1em}}
  \label{fig:factor_graph_s_p_r_1}
\end{figure}

We want to estimate the posterior probability $p(\calX_t|\mathcal{Z}_t)$ of our variables $\calX_t$ (\cref{eq:state_vector}) using the set of measurements $\mathcal{Z}_t$ (\cref{eq:measurements}).
Using standard independence assumptions between measurements, we arrive to the following formulation where we grouped the different terms in factors $\phi$:
\begin{subequations}\label{eq:factor_form}
  \begin{align}
    &p(\calX_t|\mathcal{Z}_t) \overset{(a)}{\propto} p(\calX_t)p(\mathcal{Z}_t|\calX_t) \nonumber\\
    &= \phi_0(\mathbf{x}_0)\prod_{l_c\in\Lambda_t}\prod_{\pi\in\Pi_t}\!\phi_{R}(\bm{\rho}_{l_c}, \bm\pi_\pi)^{\delta(l_c, \pi)} \label{eq:factor_form_a}\\
    &\quad\quad\quad\quad\prod_{(i,j)\in\calK_t}\!\!\!\phi_{\text{IMU}}(\mathbf{x}_i, \mathbf{x}_j) \label{eq:factor_form_b}\\
    &\prod_{i\in\calK_t}\prod_{l_c\in\Lambda_t(i)}\!\phi_{l_c}(\mathbf{x}_i, \bm{\rho}_{l_c}) \!\!\prod_{l_s\in\calS_t}\!\phi_{l_s}(\mathbf{x}_{i\in\calK_t(l_s)}), \label{eq:factor_form_c}
  \end{align}
\end{subequations}
where we apply the Bayes rule in (a), and ignore the normalization factor since it will not influence the result (\cref{sssec:map_estimation}).
\Cref{eq:factor_form_a} corresponds to the prior information we have about $\calX_t$.
The factor $\phi_0$ represents a prior on the first state of the optimization's time-horizon.
The following terms in \cref{eq:factor_form_a} encode regularity factors $\phi_R$ between constrained landmarks $l_c$ and planes $\pi$.
We also introduce the data association term $\delta(l_c, \pi)$, which returns a value of 1 if the landmark $l_c$ is associated to the plane $\pi$, 0 otherwise (\cref{sssec:data_association}).
In \cref{eq:factor_form_b}, we have the factor corresponding to the IMU measurements which depends only on the consecutive keyframes $(i, j)\in\calK_t$.
\cref{eq:factor_form_c} encodes the factors corresponding to the camera measurements.
We add a projection factor $\phi_{l_c}$ for each observation of a constrained landmark $l_c$,
where we denote by $\Lambda_t(i) \subseteq \Lambda_t$ the set of constrained landmarks seen by keyframe $i$.
Finally, we add structureless factors $\phi_{l_s}$ for each of the landmarks $l_s \in \calS_t$;
note that these factors depend on the subset of keyframes that observe $l_s$, which we denote by $\calK_t(l_s) \subseteq \calK_t$.
In \cref{fig:factor_graph_s_p_r_1}, we use the expressiveness of factor graphs \cite{Kschischang01it, Dellaert17now} to show the dependencies between the variables in \cref{eq:factor_form}.

\subsubsection{MAP Estimation}
\label{sssec:map_estimation}
Since we are only interested in the most likely $\calX_t$ given the measurements $\mathcal{Z}_t$, we calculate the \emph{maximum a posteriori} (MAP) estimator $\calX_t^{\text{MAP}}$.
Minimizing the negative logarithm of the posterior probability in \cref{eq:factor_form} (under the assumption of zero-mean Gaussian noise) leads to a nonlinear least-squares problem:

\begin{equation*}
  \begin{split}
   &\calX_t^{\text{MAP}} = \arg\min_{\calX_t} \normsq{\mathbf{r}_0}{\Sigma_0} + \!\!\!\sum_{l_c\in\Lambda_t}\!\sum_{\pi\in\Pi_t}\delta(l_c, \pi)\normsq{\mathbf{r}_{R}}{\Sigma_{R}} \\
                  &+ \!\!\!\!\!\sum_{(i,j)\in\calK_t}\!\!\!\normsq{\mathbf{r}_{\measimu_\subimu}}{\Sigma_{ij}} \!\!\! + \!\!\! \sum_{i\in\calK_t}\sum_{l_c\in\Lambda_t(i)} \normsq{\mathbf{r}_{l_c}}{\Sigma_\mathcal{C}} \!\!\! + \!\!\! \sum_{l_s\in\calS_t} \normsq{\mathbf{r}_{l_s}}{\Sigma_\mathcal{S}}\!\!,
  \end{split}
\end{equation*}
where $\mathbf{r}$ represents the residual errors, and $\mathbf{\Sigma}$ the covariance matrices.
We refer the reader to \cite[Sec. VI, VII]{Forster17troOnmanifold} for the actual formulation of the preintegrated IMU factors $\phi_{\text{IMU}}$ and structureless factors $\phi_{l_s}$, as well as the underlying residual functions $\mathbf{r}_{\text{IMU}}$, $\mathbf{r}_{l_s}$.
For the projection factors $\phi_{l_c}$, we use a standard monocular and stereo reprojection error as in \cite{Carlone17icra-vioAttention}.

\subsubsection{Regularity Constraints}
\label{sssec:regularity_constraints}

For the regularity residuals $\bm{r}_{R}$, we use a co-planarity constraint between a landmark $\bm\rho_{l_c}\in\mathbb{R}^3$ and a plane $\bm{\pi} = \{\bm{n}, d\}$, where $\bm{n}$ is the normal of the plane, which lives in the $\Stwo\doteq\{\mathbf{n} = (n_x, n_y, n_z)^T \big| \|\mathbf{n}\| = 1\}$ manifold, and $d\in\mathbb{R}$ is the distance to the world origin:
$\textstyle\residual_{R} = \bm{n} \cdot \bm{\rho}_{l_c} - d$.
This plane representation is nevertheless an over-parametrization that will lead to a singular information matrix.
This is not amenable for Gauss-Newton optimization, since it leads to singularities in the normal equations~\cite{Kaess15icra}.
To avoid this problem, we optimize in the tangent space $T_{\Normal}S^2 \sim \Real^{2}$ %
of $S^2$ and define a suitable retraction $\calR_{\Normal}(\bm{v}): T_{\Normal}S^2 \in \mathbb{R}^2 \rightarrow \Stwo$
to map changes in the tangent space to changes of the normals in $\Stwo$~\cite{Forster17troOnmanifold}. %
In other words, we rewrite the residuals as:
\begin{equation}
  \residual_R(\bm{v},d) = \calR_{\Normal}(\bm{v})^\transpose \cdot \landmark - d
  \label{eq:coplanarity_constraint}
\end{equation}
and optimize with respect to the minimal parametrization $[\bm{v}, d] \in \Real^3$.
This is similar to \cite{Kaess15icra}, but we work on the manifold $\Stwo$ instead of adopting a quaternion parametrization.
Note that a single co-planarity constraint, as defined in \cref{eq:coplanarity_constraint}, is not sufficient to constrain a plane variable, and a minimum of three are needed instead.
Nevertheless, degenerate configurations exist, e.g.~three landmarks on a line would not fully constrain a plane.
Therefore, we ensure that a plane candidate has a minimum number of constraints before adding it to the optimization problem.

\section{Experimental Results}
\label{sec:results}

We benchmark the proposed approach against the state of the art on real datasets, and evaluate
 trajectory and map estimation accuracy, as well as runtime.
We use the \Euroc dataset~\cite{Burri15ijrr}, which contains visual and inertial data recorded from an micro aerial vehicle flying indoors.
The \Euroc dataset includes eleven datasets in total, recorded in two different scenarios.
The \textit{Machine Hall} scenario (\texttt{MH}) is the interior of an industrial facility.
It contains little (planar) regularities.
The \textit{Vicon Room} (\texttt{V}) is similar to an office room where walls, floor, and ceiling are visible, as well as other planar surfaces (boxes, stacked mattresses).

\mysubsection{Compared techniques}
To assess the advantages of our proposed approach, we compare three formulations that build one on top of another.
First, we denote as \textbf{S} the approach that uses only Structureless factors ($\phi_{l_s}$, in \cref{eq:factor_form_c}).
Second, we denote as \textbf{SP} the approach that uses Structureless factors, combined with Projection factors for those landmarks that have co-planarity constraints ($\phi_{l_c}$, in \cref{eq:factor_form_c}), but without using regularity factors.
Finally, we denote as \textbf{SPR} our proposed formulation using Structureless, Projection and Regularity factors ($\phi_{R}$, in \cref{eq:factor_form_a}).
The IMU factors ($\phi_{\text{IMU}}$, in \cref{eq:factor_form_b}) are implicitly used in all three formulations.
We also compare our results with other state-of-the-art implementations in \cref{tab:ape_accuracy_comparison_sopa}.
In particular, we compare the Root Mean Squared Error (RMSE) of our pipeline against OKVIS~\cite{Leutenegger13rss}, MSCKF~\cite{Mourikis07icra},
 ROVIO~\cite{Blosch15iros}, VINS-MONO~\cite{Qin17arxiv}, and SVO-GTSAM~\cite{Forster17troOnmanifold}, using the reported values in~\cite{Delmerico18benchmark}.
Note that these algorithms use a monocular camera, while we use a stereo camera.
Therefore, while \cite{Delmerico18benchmark} aligns the trajectories using $\mathrm{Sim}(3)$, we use instead $\mathrm{SE}(3)$.
Nevertheless, the scale is observable for all algorithms since they use an IMU.
No algorithm uses loop-closure.

\subsection{Localization Performance}
\label{ssec:state_estimation}

\mysubsection{Absolute Translation Error (ATE)}
\label{ssec:absolute_pose_error}
The ATE looks at the translational part of the relative pose between the ground truth pose and the corresponding estimated pose at a given timestamp.
We first align our estimated trajectory with the ground truth trajectory both temporally and spatially (in SE(3)), as explained in \cite[Sec. 4.2.1]{RosinolMT}.
We refrain from using the rotational part since the trajectory alignment ignores the orientation of the pose estimates.
\Cref{tab:ape_all_datasets_pipelines} shows the ATE for the pipelines S, SP, and our proposed approach SPR on the \Euroc dataset.

First, if we look at the performance of the different algorithmic variants for the datasets \texttt{MH\_03}, \texttt{MH\_04} and \texttt{MH\_05} in \cref{tab:ape_all_datasets_pipelines}, we observe that all methods perform equally.
This is because in these datasets no structural regularities were detected.
Hence, the pipelines SP and SPR gracefully downgrade to a standard structureless VIO pipeline (S).
Second, looking at the results
 for dataset \texttt{V2\_03}, we observe that both the SP and the SPR pipelines achieve the exact same performance.
In this case, structural regularities are detected, resulting in Projection factors being used.
Nevertheless, since the number of regularities detected is not sufficient to spawn a new plane estimate,
no structural regularities are actually enforced.
Finally, \cref{tab:ape_all_datasets_pipelines} shows that the SPR pipeline consistently achieves better results over the rest of datasets where structural regularities are detected and enforced.
In particular, SPR decreases the median APE by 27.6\% compared to the SP approach for dataset \texttt{V1\_02}, which has multiple planes.

\Cref{tab:ape_accuracy_comparison_sopa} shows that the SPR approach achieves the best results when compared with the state-of-the-art
 on datasets with structural regularities, such as in datasets \texttt{V1\_01} and \texttt{V1\_02},
  where multiple planes are present (walls, floor).
    We observe a $19\%$ improvement compared to the next best performing algorithm (SVO-GTSAM) in dataset \texttt{V1\_01},
     and a $26\%$ improvement in dataset \texttt{V1\_02} compared to ROVIO and VINS-MONO, which achieve the next best results.
    We also see that the performance of our pipeline is on-par with other state-of-the-art approaches when no structural regularities are present, such as in datasets \texttt{MH\_04} and \texttt{MH\_05}.

\begin{table}[t!]
  \centering
  \caption{ATE for pipelines S, SP, and SPR. Our proposed approach SPR achieves the best results for all datasets where structural regularities are detected and enforced.}
  \label{tab:ape_all_datasets_pipelines}
  \begin{tabularx}{\columnwidth}{l *6{Y}}
    \toprule
    & \multicolumn{6}{c}{ATE [cm]} \\
    \cmidrule{2-7}
    & \multicolumn{2}{c}{S} & \multicolumn{2}{c}{SP} & \multicolumn{2}{c}{SPR (\textbf{Proposed})} \\
    \cmidrule(r){2-3} \cmidrule(){4-5} \cmidrule(l){6-7}
    EuRoC Sequence & Median & RMSE & Median & RMSE & Median & RMSE \\
    \midrule
             MH\_01\_easy & 13.7 & 15.0 & 12.4 & 15.0 & \textbf{{10.7}} & \textbf{{14.5}} \\
             MH\_02\_easy & 12.9 & 13.1 & 17.6 & 16.7 & \textbf{{12.6}} & \textbf{{13.0}} \\
             MH\_03\_medium & \textbf{21.0} & \textbf{21.2} & \textbf{21.0} & \textbf{21.2} &\textbf{21.0} & \textbf{21.2} \\
             MH\_04\_difficult & \textbf{17.3} & \textbf{21.7} & \textbf{17.3} & \textbf{21.7} & \textbf{17.3} &  \textbf{21.7} \\
             MH\_05\_difficult & \textbf{21.6} & \textbf{22.6} & \textbf{21.6} & \textbf{22.6} & \textbf{21.6} &  \textbf{22.6} \\
             V1\_01\_easy & 5.6 & 6.4 & 6.2 & 7.7 & \textbf{{5.3}} & \textbf{{5.7}} \\
           V1\_02\_medium & 7.7 & 8.9 & 8.7 & 9.4 & \textbf{{6.3}} & \textbf{{7.4}} \\
        V1\_03\_difficult & 17.7 & 23.1 & 13.6 & 17.6 & \textbf{{13.5}} & \textbf{{16.7}} \\
             V2\_01\_easy & 8.0 & 8.9 & 6.6 & 8.2 & \textbf{{6.3}} & \textbf{{8.1}} \\
           V2\_02\_medium & 8.8 & 12.7 & 9.1 & 13.5 & \textbf{{7.1}} & \textbf{{10.3}} \\
           V2\_03\_difficult & 37.9 & 41.5 & \textbf{26.0} & \textbf{27.2} & \textbf{26.0} & \textbf{27.2} \\
    \bottomrule
  \end{tabularx}%
\end{table}
 
\begin{table}[t!]
  \centering
  \caption{RMSE of the state-of-the-art techniques (reported values from \cite{Delmerico18benchmark}) compared to our proposed SPR pipeline, on the \Euroc dataset. A cross ($\times$) states that the pipeline failed.
  In \textbf{bold} the best result, in \textcolor{blue}{blue} the second best.}
  \label{tab:ape_accuracy_comparison_sopa}
  \begin{tabularx}{\columnwidth}{l *6{Y}}
    \toprule
    & \multicolumn{6}{c}{RMSE ATE [cm]} \\
    \cmidrule(l){2-7}
    Sequence  & OKVIS & MSCKF & ROVIO & VINS-MONO & SVO-GTSAM & \textbf{SPR} \\
    \midrule
    MH\_01 & 16 & 42 & 21 & 27 & \textbf{5} & \textbf{\textcolor{blue}{14}} \\
    MH\_02 & 22 & 45 & 25 & \textbf{\textcolor{blue}{12}} & \textbf{3} & 13 \\
    MH\_03 & 24 & 23 & 25 & \textbf{\textcolor{blue}{13}} & \textbf{12} & 21 \\
    MH\_04 & 34 & 37 & 49 & 23 & \textbf{13} & \textbf{\textcolor{blue}{22}} \\
    MH\_05 & 47 & 48 & 52 & 35 & \textbf{16} & \textbf{\textcolor{blue}{23}} \\
    V1\_01 & 9 & 34 & 10 & \textbf{\textcolor{blue}{7}} & \textbf{\textcolor{blue}{7}} & \textbf{6} \\
    V1\_02 & 20 & 20 & \textbf{\textcolor{blue}{10}} & \textbf{\textcolor{blue}{10}} & 11 & \textbf{7} \\
    V1\_03 & 24 & 67 & \textbf{\textcolor{blue}{14}} & \textbf{13} & $\times$ & 17 \\
    V2\_01 & 13 & 10 & 12 & \textbf{\textcolor{blue}{8}} & \textbf{7} & \textbf{\textcolor{blue}{8}} \\
    V2\_02 & 16 & 16 & 14 & \textbf{8} & $\times$ & \textbf{\textcolor{blue}{10}} \\
    V2\_03 & 29 & 113 & \textbf{14} & \textbf{\textcolor{blue}{21}} & $\times$ & 27 \\
    \bottomrule
  \end{tabularx}
  \vspace{-2.10em}
\end{table}

\mysubsection{Relative Pose Error (RPE)}
\label{ssec:relative_pose_error}
While the ATE provides information on the global consistency of the trajectory estimate, it does not provide insights on the moment in time when the erroneous estimates happen.
Instead, RPE is a metric for investigating the local consistency of a trajectory.
RPE aligns the estimated and ground truth pose for a given frame $i$, and then computes the error of the estimated pose for a frame $j>i$ at a fixed distance farther along the trajectory.
We calculate the RPE from frame $i$ to $j$ in translation and rotation (absolute angular error) \cite[Sec. 4.2.3]{RosinolMT}.
As \cite{Geiger12cvpr}, we evaluate the RPE over all possible trajectories of a given length, and for different lengths.

\begin{figure}[tbp]
  \centering     %
  \includegraphics[trim={0.1cm 0 0cm 0.2cm},clip,width=0.8\columnwidth]{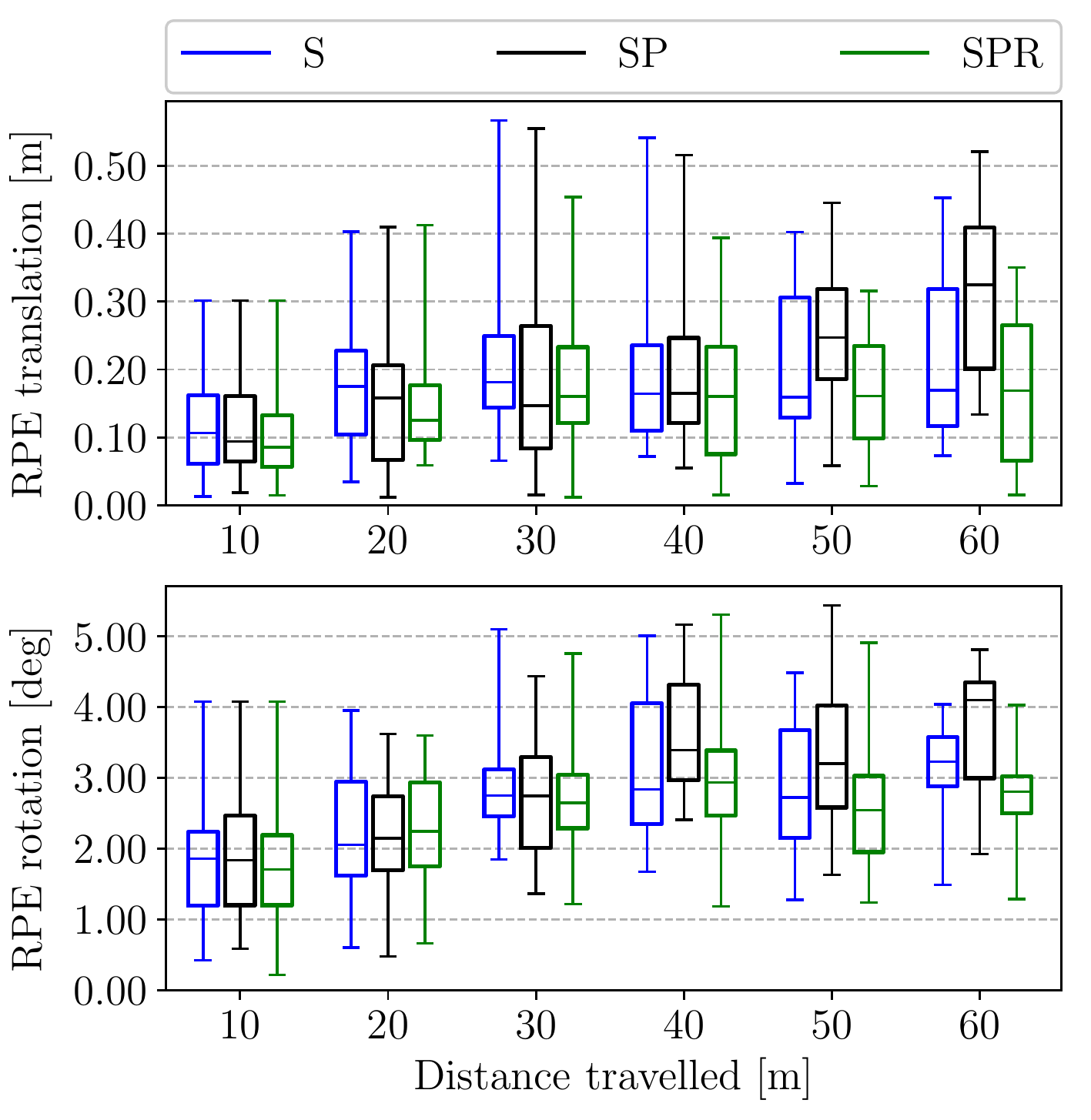}
  \caption{Boxplots of the RPE on dataset \texttt{V2\_02} for the approaches S, SP, and SPR (proposed). \vspace{-1em}}
  \label{fig:boxplot_rpe}
\end{figure}

\Cref{fig:boxplot_rpe} shows the results for dataset \texttt{V2\_02}, where we observe that using our proposed pipeline SPR,
 with respect to the SP pipeline, leads to: (i) an average improvement of the median of the RPE over all trajectory lengths of 20\% in translation and 15\% in rotation,
  and (ii) a maximum accuracy improvement of 50\% in translation and 30\% in rotation of the median of the RPE.

\subsection{Mapping quality}
\label{ssec:mapping_quality}
\begin{figure}[tb]
  \centering     %
  \includegraphics[height=0.5\columnwidth, trim={0 1.5cm 0 0cm},clip,width=\columnwidth]{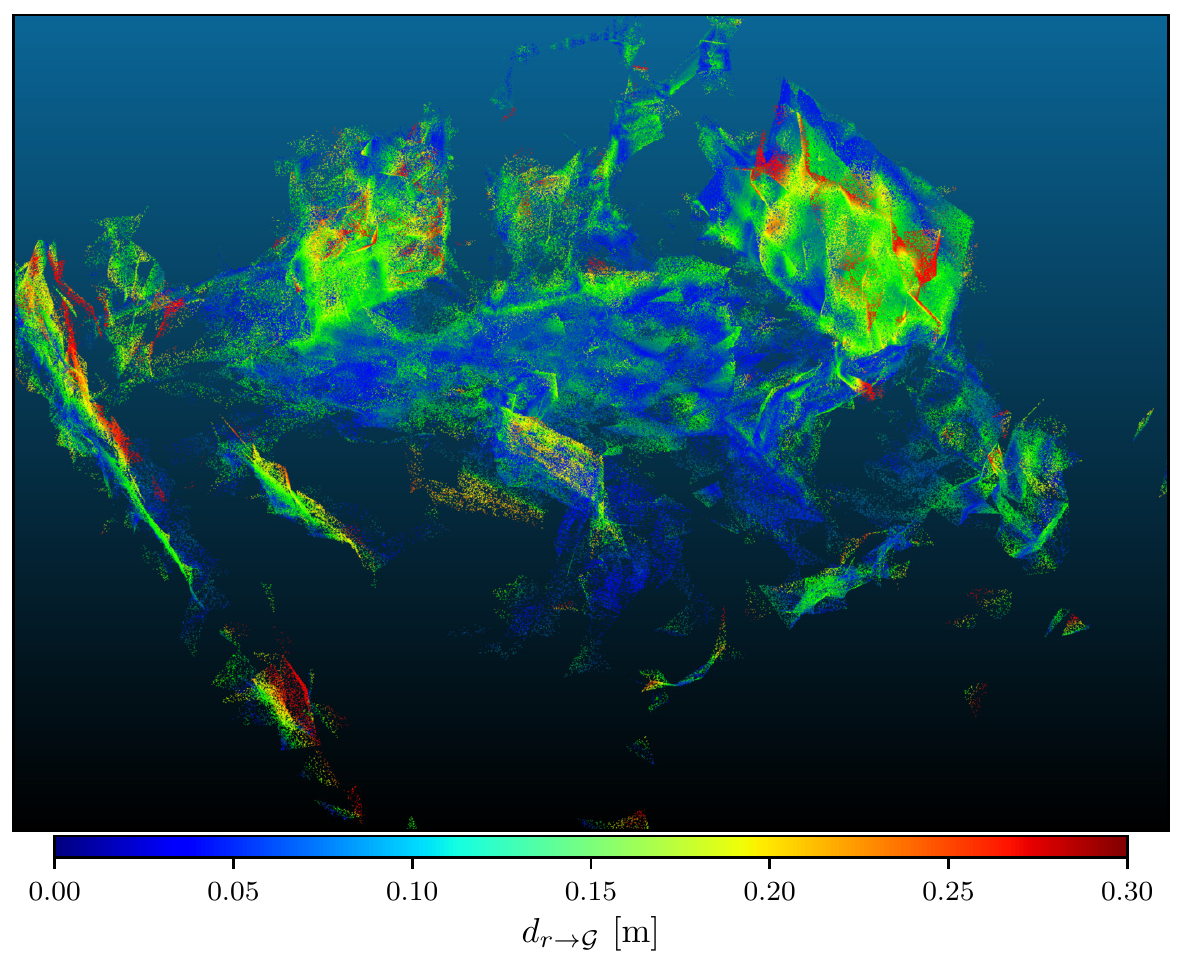}
  \includegraphics[height=0.5\columnwidth, trim={0 1.5cm 0 0cm},clip,width=\columnwidth]{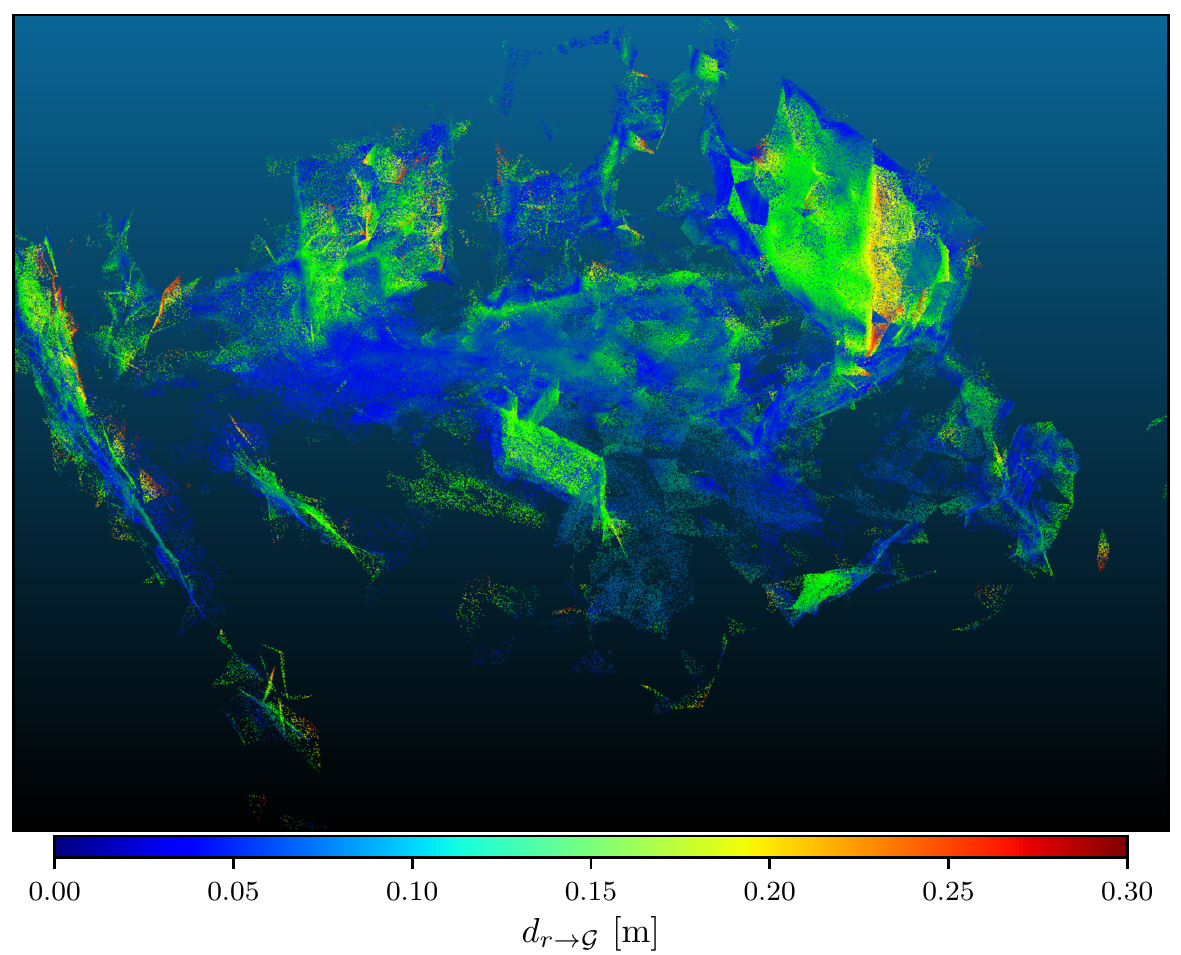}
  \includegraphics[trim={0 0 0 8.425cm},clip,width=\columnwidth]{results/S_P_R_Mesh/frames_animation/frame_000000_S_P_R-eps-converted-to.pdf}

  \caption{Point cloud sampled from the estimated 3D mesh color-encoded with the distance to the ground truth point cloud (\texttt{V1\_01}), for SP approach (top) and SPR (bottom).
   \vspace{-1em}}
  \label{fig:accuracy_mesh}
\end{figure}

We use the ground truth point cloud for dataset \texttt{V1} to assess the quality of the 3D mesh by calculating its \textit{accuracy}, as defined in \cite{Schoeps2017cvpr}.
To compare the mesh with the ground truth point cloud, we compute a point cloud by sampling the mesh with a uniform density of $10^3~\text{points}/m^2$.
We also register the resulting point cloud to the ground truth point cloud.
  In \cref{fig:accuracy_mesh}, we color-encode each point $r$ on the estimated point cloud with its distance to the closest point in the ground-truth point cloud $\mathcal{G}$ ($d_{r \to \mathcal{G}}$).
  We can observe that, when we do not enforce structural regularities, significant errors are actually present on the planar surfaces, especially on the walls (\cref{fig:accuracy_mesh} top).
  Instead, when regularities are enforced, the errors on the walls and the floor are reduced (\cref{fig:accuracy_mesh} bottom).
A closer view on the wall itself (\cref{fig:intro}(c)-(d)) provides an illustrative example of how adding co-planarity constraints results in smoother walls.

\subsection{Timing}
\label{ssec:timing}

The pipelines S, SP, and SPR differ in that they try to solve an increasingly complicated optimization problem.
While the S pipeline does not include neither the 3D landmarks nor the planes as variables in the optimization problem, the SP pipeline includes 3D landmarks, and the pipeline using regularities (SPR) further includes planes as variables.
Moreover, SP has significantly less factors between variables than the SPR pipeline.
Hence, we can expect that the optimization times for the different pipelines will be each bounded by the other as $t_{S}^{opt} < t_{SP}^{opt} < t_{SPR}^{opt}$, where $t_{X}^{opt}$ is the time taken to solve the optimization problem of pipeline X.

\begin{figure}[tb]
  \centering     %
    \resizebox{\columnwidth}{!}{\input{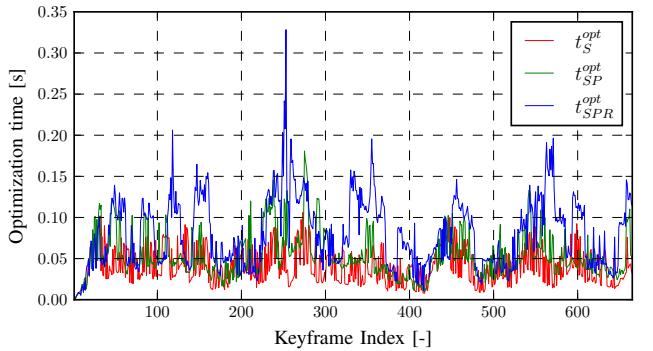}}
  \caption{Comparison of the time to solve the optimization problem for pipeline S, SP, and SPR for dataset \texttt{V1\_01}. \vspace{-1em}}
  \label{fig:optimization_time}
\end{figure}

\Cref{fig:optimization_time} shows the time taken to solve the optimization problem for each type of pipeline.
We observe that the optimization time follows roughly the expected distribution.
We also notice that if the number of plane variables is large ($\sim 10^1$), and consequently the number of constraints between landmarks and planes also gets large ($\sim 10^2$), the optimization problem cannot be solved in real-time
(see keyframe index 250 in \cref{fig:optimization_time}). This can be avoided by restricting the number of planes in the optimization.
Finally, the SPR pipeline has the overhead of generating the mesh. Nevertheless, it takes just $8$ms per frame.

\section{Conclusion}
\label{sec:conclusions}

We present a VIO algorithm capable of incrementally building a 3D mesh of the scene restricted to a receding time-horizon.
Moreover, we show that we can improve the state estimation and mesh by enforcing structural regularities present in the scene.
Hence, we provide a tightly coupled approach to regularize the mesh and improve the state estimates simultaneously.

Finally, while the results presented are promising, we are not yet enforcing higher level regularities (such as parallelism or orthogonality) between planes.
Therefore, these improvements could be even larger, potentially rivaling pipelines enforcing loop-closures.

\IEEEtriggeratref{14}
\bibliographystyle{IEEEtran} %

\end{document}